\documentclass{article}
\usepackage{spconf,amsmath,graphicx}
\usepackage{amssymb}
\usepackage{comment}
\usepackage{algorithm}
\usepackage[noend]{algpseudocode}
\usepackage{url}
\usepackage[colorinlistoftodos]{todonotes}

\newcommand{\mycomment}[1]{}

\def\x{{\mathbf x}}
\def\z{{\mathbf z}}
\def\r{{\mathbf r}}
\def\1{{\mathbf 1}}
\def\0{{\mathbf 0}}

\def\D{{\cal D}}
\def\M{{\cal M}}
\def\N{{\cal N}}
\title{How to train your VAE}
\name{{Mariano Rivera}
\thanks{This Work has been submitted to the IEEE for possible publication. Copyright may be transferred without notice, after which this version may no longer be accessible.}}
\address{Centro de Investigacion en Matematicas A.C. \\
         Guanajuato, Gto. 36120, Mexico}
\begin{document}
%
\maketitle

\begin{abstract}

Variational Autoencoders (VAEs) have become a cornerstone in generative modeling and representation learning within machine learning. This paper explores a nuanced aspect of VAEs, focusing on interpreting the Kullback–Leibler (KL) Divergence, a critical component within the Evidence Lower Bound (ELBO) that governs the trade-off between reconstruction accuracy and regularization.
Meanwhile, the KL Divergence enforces alignment between latent variable distributions and a prior imposing a structure on the overall latent space but leaves individual variable distributions unconstrained. The proposed method redefines the ELBO with a mixture of Gaussians for the posterior probability, introduces a regularization term to prevent variance collapse, and employs a PatchGAN discriminator to enhance texture realism. Implementation details involve ResNetV2 architectures for both the Encoder and Decoder. The experiments demonstrate the ability to generate realistic faces, offering a promising solution for enhancing VAE-based generative models.
\end{abstract}


\section{Introduction}
\label{sec:intro}

\noindent Variational Autoencoders (VAEs) have emerged as a robust framework for generative modeling and representation learning in machine learning \cite{kingma2013auto,rezende2014stochastic}. Although more complex variations of VAEs have been proposed (e.g., Hierarchical VAEs \cite{havtorn2021hierarchical} Wasserstein-AE \cite{tolstikhin2018wasserstein}, Beta-VAE \cite{betaVAE} and VQ-VAEs \cite{razavi2019generating}), the original VAEs continue to be of great interest due to their elegant theoretical formulation. In the pursuit of understanding and enhancing the capabilities of VAEs, this paper presents a variant of a fundamental component: the Evidence Lower Bound (ELBO). The ELBO is a critical objective function for training VAEs, encapsulating the trade-off between reconstruction accuracy and regularization. Our focus in this work centers on a detailed analysis of the interpretation of the regularization term within the ELBO, known as the Kullback--Leibler (KL) Divergence \cite{kingma2013auto,rezende2014stochastic,rezende2015variational}.

The KL Divergence has a pivotal role in VAEs as it enforces alignment between the distribution of latent variables and a prior distribution, often assumed to be a Normal distribution with zero mean and unit covariance. This divergence term promotes the emergence of structured latent representations, a cornerstone of effective generative models. However, our research delves deeper into the latent spatial structure. We note that the KL Divergence does not impose any constraints on the individual distributions of the latent variable associated with a datum (individual posterior);  this has implications for the diversity of latent vector behaviors in VAEs. In light of these findings, our proposal reformulates the role of KL Divergence with implications for latent space modeling. Our proposal addresses a well-known challenge in VAEs: the posterior collapse phenomenon \cite{he2019lagging}, which can lead to a loss of distinctiveness in generated data. Specifically, we demonstrate the successful application of our approach in generating realistic facial images while retaining crucial variations in features like hairstyles and specific facial attributes \cite{higgins2017beta,razavi2019generating}

The main contribution in this work is an ELBO that reduces the collapse of the posterior towards the anterior (observed as the generation of very similar, blurry images). Our ELBO can be seen as an opposite version of Beta-VAE that preferentially weights quality reconstructions and a regularization term that prevents the collapse of individual posteriors to a Dirac delta. Additionally, we use a VAE-GAN type scheme that promotes realistic data. Finally, we employ a Residual type architecture for the encoder and decoder to increase the network capabilities. The main contributions affect the training process, so our results must be compared directly with those of a standard VAE \cite{kingma2013auto,rezende2014stochastic} or Beta-VAE \cite{betaVAE}. Qualitative comparisons of the presented results versus those reported in the classic VAE or Beta-VAE formulations show the superior performance of our proposal.

\begin{figure}[!t]
\centering
\includegraphics[width=0.9\linewidth]{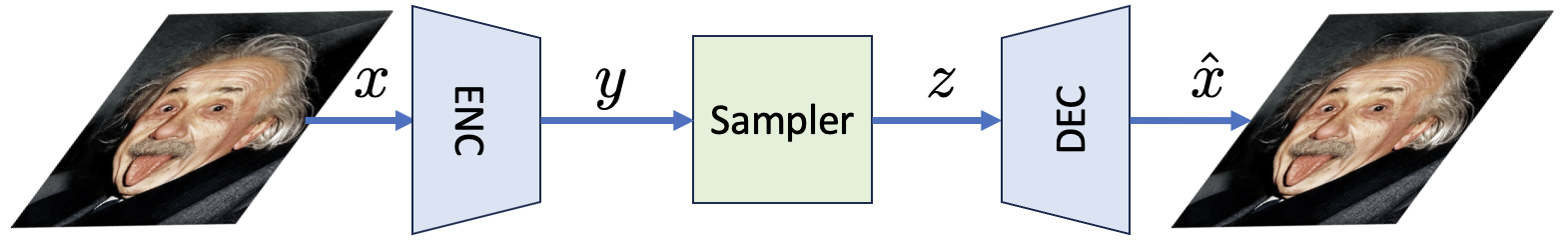}
\caption{General scheme of a VAE} 
\label{fig:vae}
\end{figure}

We organize this paper as follow. We present in Section \ref{sec:background} a brief review of the standard VAE model. Then, in section \ref{sec:proposal}, we present a mixture of Gaussian's for defining the posterior probability $q({\bf z} | {\bf x})$ in the VAE, which leads us to redefine the ELBO, and we present practical implementation steps. Thus, we offer experimental results that support our claims in section \ref{sec:experiments}. By advancing our understanding of the regularization mechanism in VAEs and its practical implications, we contribute to the ongoing pursuit of enhancing generative modeling techniques for complex data distributions. Through this work, we hope to shed light on the interplay between regularization and diversity in latent spatial representations, contributing to advancing generative models in machine learning and artificial intelligence.
The code that implements the proposal is available in \url{https://github.com/marianorivera/How2TrainUrVAE}.

\section{Background}
\label{sec:background}

Variational Autoencoder (VAE) is an extension of the traditional Autoencoder (AE) concept \cite{kingma2013auto,rezende2014stochastic}. The pivotal breakthrough of the VAE lies in its novel approach to shaping the distribution of latent variables, urging them to adhere to a simplified parametric distribution. This streamlining eases the generation of latent variables. Hence, it also facilitates the creation of complex data instances, as demonstrated by the case of faces—the example we utilize to illustrate our proposition.

The goal is to learn the generative model of ${\mathcal D}$ a dataset with $N \ge 1$ \emph{independent and identically distributed} (iid) data points: ${\mathcal D} = \{ {\bf x}^{(i)}\}_{i=1}^N$ with ${\bf x}^{(i)} \in \mathbb{R}^m$; assuming a set of latent variables, one for each data point: ${\mathcal Z} = \{ {\bf z}^{(i)}\}_{i=1}^N$ with  ${\bf z}^{(i)} \in \mathbb{R}^d$, and $d < m$. In this way, we have $({\bf x}, {\bf z}) \sim p({\bf x}, {\bf z}) = p({\bf x} | {\bf z}) p({\bf z})$, which suggests a two-step generative model: 
\begin{enumerate}
    \item Generation of a random latent variable ${\bf z} \sim p({\bf z})$; 
    
    \item Generation of the data point ${\bf x} \sim p({\bf x} | {\bf z})$. 
\end{enumerate}
The latent variables' distribution $p({\bf z})$ (prior) can be assumed. Then, one must estimate the conditional distribution $p({\bf x} | {\bf z})$ (likelihood). The problem arises because only the data points, ${\mathcal D}$, are observed, while the latent variables, ${\mathcal Z}$, remain hidden. That is, one does not know the pairs $({\bf x}^{(i)}, {\bf z}^{(i)})$. To determine the ${\bf z}^{(i)}$ corresponding to ${\bf x}^{(i)}$, it is necessary to estimate the conditional distribution $p( {\bf z} | {\bf x})$. Using Bayes' rule, one can write the posterior as
\begin{equation}
    p( {\bf z} | {\bf x}) =  p({\bf x} | {\bf z}) p({\bf z}) / p({\bf x}).
\end{equation}
However, calculating the evidence $p({\bf x}) $ is an intractable problem \cite{kingma2013auto,rezende2014stochastic}. Therefore, instead of calculating the true posterior $p( {\bf z} | {\bf x})$ one computes an approximation $q( {\bf z} | {\bf x})$. A VAE is an autoencoder where: $q_\phi( {\bf z} | {\bf x})$ is a neural network that encodes the data ${\bf x}^{(i)}$ into latent variables ${\bf z}^{(i)}$, and $p_\theta({\bf x} | {\bf z})$ decodes the latent variable ${\bf z}^{(i)}$ into ${\bf x}^{(i)}$; where, $\phi$ and $\theta$ are the parameters of the encoder and decoder, respectively. 
This stochastic generation process can be represented as:
\begin{equation}
  {\bf x}^{(i)} \overset{q_\phi} {\leadsto} {\bf z}^{(i)} \overset{p_\theta} {\mapsto}\hat {\bf x}^{(i)},
\end{equation}
where $\hat {\bf x}^{(i)}$ is the generated data similar to ${\bf x}^{(i)}$. The Encoder is stochastic, meaning that if we encode $K$ times the input ${\bf x}^{(i)}$, we obtain $K$ distinct values of the latent variable: $Z^{(i)}=\{{\bf z}^{(i,k)}\}_{k=1}^K$. 
Then, the VAE's training is performed by minimizing a cost function called the ELBO (Evidence Lower Bound):
\begin{align}
\mathrm{ELBO}(\phi, \theta) 
= & \beta_1 \,\mathrm{KL}\bigl( q_\phi(\mathbf{z}|\mathbf{x}) || p_\lambda(\mathbf{z}) \bigr)
\label{eq:ELBO_KL}\\
& - \beta_ 2\, \mathbb{E}_{q_\phi(\mathbf{z}|\mathbf{x})} \left[ \log p_\theta(\mathbf{x}|\mathbf{z}) \right]
\label{eq:ELBO_l}
\end{align}
where the parameters $\beta_1$ and $\beta_2$ weigh the relative contribution of each term to the total cost. The standard VAE formulation uses $\beta_1=\beta_2=1$, meanwhile $\beta$-VAE propose to use $\beta_1>1$ and $\beta_2=1$ to improve interpretable factorised latent representations; \emph{i.e.}, to learn disentangled latents \cite{higgins2017beta}. The second term \eqref{eq:ELBO_l} is the negative log-likelihood that enforces the generated data, $\hat{\bf x}^{(i)}$, to be likely to belong to the training dataset. The first term \eqref{eq:ELBO_KL} is the Kullback-Leibler divergence that enforces the approximated posterior $q_{\phi}$ to be similar to a chosen prior $p_{\lambda}$; \emph{i.e.} ${\bf z} \sim p_\lambda({\bf z})$ and $\lambda$ are the parameters of the prior. The most common approach is to assume the prior distribution $p_\lambda({\bf z})$ a multivariate Gaussian with a mean of zero and an identity covariance matrix: 
\begin{equation}
p_\lambda(\mathbf{z}) = \mathcal{N}(0,I).
\end{equation}
Since neural networks are deterministic, in order to define $q_\phi(\mathbf{z}|\mathbf{x})$, the stochastic behavior can implemented using the reparametrization trick:
\begin{equation}
{\bf z}^{(i,k)}_j = {\bf \mu}^{(i)}_j + \epsilon^{(k)}_j {\bf \sigma}^{(i)}_j, \;\; \text{for} \;\; j=1,2,\ldots, d;
\end{equation}
where $\epsilon \sim \mathcal{N}(0,I)$ and $\phi^{(i)} = ({\bf \mu}^{(i)}, {\bf \sigma}^{(i)})$ are the (deterministic) vectors that the network encodes the data ${\bf x}^{(i)}$ into. 
\begin{figure}[!t]
    \centering
    \includegraphics[width=0.4\textwidth]{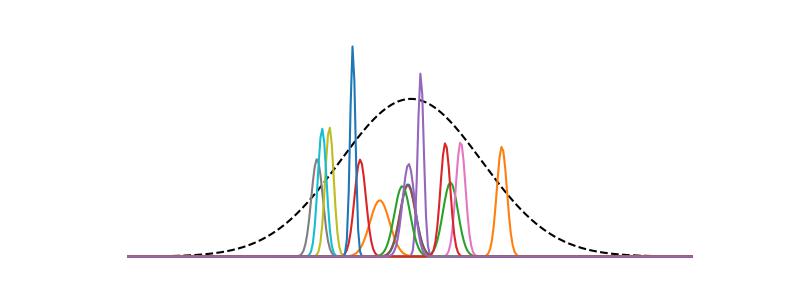}
    \caption{Illustration of the global posterior $q_{\phi}$  as a product of individual posteriors $q_{\phi_i}$.}
    \label{fig:posterior}
\end{figure}
Hence, one can factorize the posterior as  \cite{kingma2019introduction}:
\begin{align}
    \label{eq:prodGss}
    q_\phi({\bf z}|{\bf x}) &= \prod_{{\bf x}^{(i)} \in \mathcal{D}} q_\phi({\bf z} | {\bf x}={\bf x}^{(i)}) \\
    &= \prod_{{\bf x}^{(i)} \in \mathcal{D}}\mathcal{N}({\bf z}; \mu^{(i)}, \sigma^{(i)}).
    \label{eq:prodNorm}
    \\ &= \prod_{{\bf x}^{(i)} \in \mathcal{D}} \prod_{j=1}^{d} \mathcal{N}({\bf z}_j; \mu_j^{(i)}, \sigma_j^{(i)}).
\end{align}
This factorization takes advantage of the fact that the product of Gaussian (densities) is a Gaussian (density).  The second row is the product of the individual posteriors (associated with each data point ${\bf x}^i$), and the third row makes explicit the marginals (of each latent dimension ${\bf x}_j$).    
Fig. \ref{fig:posterior} illustrates the Gaussian mixture model for the posterior. The dashed line represents the posterior $q_\phi$; \emph{i.e.}, the product of the individual Gaussian $q_{\phi}^{(i)}$ associated with each data point (in color). Note that while $q_\phi$ is enforced to be similar to the prior $p_{\lambda}$: a Gaussian with mean equal to zero and unitary variance. 
Since the KL between two univariate Gaussian is given by
\begin{multline}
\mathrm{KL}\bigl( \mathcal{N}(\mu_1, \sigma_1^2)) \| \mathcal{N}(\mu,_2 \sigma_2^2) \bigr) \\
        =\frac{1}{2} \left[ \frac{\sigma_1^2}{\sigma_2^2} + (\mu_1-\mu_2)^2 - 1 - \log \frac{\sigma_1^2}{\sigma_2^2} \right].
\label{eq:KLgg}
\end{multline}
Thus, according with Ref. \cite{kingma2013auto}, the KL divergence term in the ELBO \eqref{eq:ELBO_KL} takes the form:
\begin{multline}
\mathrm{KL}\bigl( q_\phi(\mathbf{z}|\mathbf{x}) || p_\lambda(\mathbf{z}) \bigr)
   = \sum_{j=1}^d  \mathrm{KL}\bigl( \mathcal{N}(\bar \mu_j, \bar \sigma_j^2) \| \mathcal{N}(0,I) \bigr) 
   \\
        =\frac{1}{2} \sum_{j=1}^d  \sum_{i: {\bf x}^{(i)} \in D} \left[ (\sigma_j^{(i)})^2 + (\mu^{(i)}_j)^2 - 1 - \log 
        (\sigma_j^{(i)})^2\right];
\label{eq:KL_king}
\end{multline}
where $\bar\mu$ is the mean vector. The ELBO minimization is efficiently achieved by using stochastic gradient descent by using samples of the training set (mini-batches): $\mathcal{M} \subset \mathcal{D}$.

\section{Method}
\label{sec:proposal}

\subsection{The posterior collapse problem and over--smoothed reconstructions}
\label{ssec:meanvarianceposterior}

VAEs often suffer from the \emph{posterior collapse} problem \cite{wang2021posterior} (latent variable collapse or KL vanishing); such a problem occurs when $\mathrm{KL}\bigl( q_\phi(\mathbf{z}|\mathbf{x}) || p_\lambda(\mathbf{z}) \bigr) \rightarrow 0$. The posterior collapse to the prior results in non-identifiable latent variables: the likelihood function is independent of the latent representation, producing a smoothness of the reconstructed images. 
To alleviate such problem there has been suggested many possible solutions, see Ref. \cite{TAKIDA2022137} and references therein. Despite many efforts to alleviate the posterior collapse, one can yet observe over-smoothed predictions \cite{betaVAE, wang2021posterior, TAKIDA2022137}. It seems that only hierarchical computational expensive VAE architectures can generate realistic textured data \cite{NEURIPS2019_9bdb8b1f,havtorn2021hierarchical}.

To reduce the blurring, a naive strategy could be to reduce the parameter $\beta_1<1$ in \eqref{eq:ELBO_KL} and pass the weight to the log-likelihood. This presents another problem that we will analyze following. Since the covariance $\bar \sigma$ of the posterior $q_\phi(\z | \x)$ is a diagonal (non--correlated) matrix, the product in \eqref{eq:prodNorm} can be decomposed in products of univariate Gaussians, one product for each dimension $j$. Thus, the $j$--th product results in
\begin{equation}
    \label{eq:prod_n_unigss}
    \prod_{i=1}^N \N (\mu_j^{(i)}, \sigma_j^{(i)}) = s_j \, \N (\bar \mu_j, \bar \sigma_j)
\end{equation}
where 
\begin{align}
    \frac{1}{(\bar \sigma_j )^2 }&=   \sum_{i=1}^N \frac{1}{(\sigma^{(i)}_j)^2}, \\
    \bar \mu_j    & = ( \bar \sigma_j)^2 \sum_i^N \frac{\mu^{(i)}_j}{(\sigma^{(i)}_j)^2},
    \label{eq:barmu}
\end{align}
and $s_j$ is a scale factor whose exact form can be found in \cite{bromiley2003products}. 
If to obtain reconstructions that are more faithful to the data, we weight more the log--likelihood term into the ELBO, then the individual posterior $q_{\phi}^{(i)}$ variance is reduced so the variations of the predictions are limited such that the VAE conducts close to a simple Autoencoder. In such case, Equations \eqref{eq:prod_n_unigss}--\eqref{eq:barmu} make evident a problem that arises when the individual posterior $q_{\phi}^{(i)}$ approaches a Dirac delta: $\sigma_j^{(i)} \approx 0$. Thus, the $j$--th entry of the posterior $q_\phi$ [see \eqref{eq:prodGss}] approaches a Dirac Delta: $\bar \sigma_j \approx 0$. As a result, the $z_j$ latent variable is meaningless. Therefore, a significantly large $\beta_1$ value would be expected to result in a blurrier reconstruction.

\subsection{The posterior as a mixture of Gaussians}
\label{ssec:posterior}

To introduce our proposal, we first note that the latent variable ${\bf z}^{(i)}$, associated with the data point ${\bf x}^{(i)}$, is also a stochastic variable with its particular distribution, the individual  posterior: 
$$
{\bf z}^{(i)} \sim q_{\phi}( {\bf z} | {\bf x}^{(i)} ) =  q^{(i)}_{\phi}( {\bf z}).
$$ 
where $q^{(i)}_{\phi}( {\bf z}) = \mathcal{N}(\mu^{(i)}, \sigma^{(i)})$ with $\sigma^{(i)}\succ 0$. Such individual posteriors need to be combined to define the posterior. A direct possibility is factorization \eqref{eq:prodGss} but others factorization are possible. In this work, instead of product \eqref{eq:prodGss}, we propose the mixture model:\begin{equation}
q_\phi(\mathbf{z} | \mathbf{x} ) = \sum_{{\bf x}^{(i)} \in \mathcal{D}} \alpha_i \, q_{\phi}( {\bf z} | {\bf x}^{(i)} );
\end{equation}
with $\alpha_i = 1/N$ (with $N = \sharp \mathcal{D}$), which implies that all data points are equally relevant. Hence, the mean of the mixture is given by:
\begin{equation}
\bar \mu = \mathbb{E} \left[q_\phi(\mathbf{z} | \mathbf{x} )\right] = \sum_i \alpha_i \mathbb{E}\left[q_{\phi_i}( \z| \x^{(i)})\right] = \mathbb{E}\left[ {\bf \mu} ^{(i)}\right],
\end{equation}
and the variance of the posterior:
\begin{equation}
    \bar \sigma = \mathrm{var} \left[q_\phi(\z | \x )\right] = \mathrm{var} \left[  \z | \x \right].
\end{equation}

Thus, we can estimate the mean and variance of the global posterior with a Stochastic Gradient strategy using data batches $\M \subset \D$.:
\begin{align}
    \label{eq:mu_bar2}
    \bar\mu_j      & \approx  \frac{1}{\sharp \mathcal{M}} \sum_{i:\x^{(i)} \in \M}  \mu^{(i)}_{j}, \\
    \label{eq:sigma2_bar2}
    \bar\sigma^2_j & \approx  \frac{1}{\sharp \mathcal{M}} \sum_{i:\x^{(i)} \in \M}  \left( \z^{(i)}_{j} - \bar\mu_j \right)^2.
\end{align}
This way of estimating the posterior's mean and variance distinguishes our proposal from the standard formulation.

\subsection{Loss and training strategy}
\label{ssec:loss}

In this section, we present our loss function. First, we note that the classic $\mathrm{KL}$ formulation \cite{kingma2013auto} encourages the individual means $\mu^{(i)}$'s to be zero, contributing to the so-called posterior collapse effect: the $\z^{(i)}$ are non--identifiable. Differently, in our proposal, we encourage the mean of those $\mu^{(i)}$'s to approach zero: 
\begin{multline}
\label{eq:KL_G}
    \mathrm{KL}_G \bigl( q_\phi(\mathbf{z}|\mathbf{x}) \, \| \, p_\lambda(\mathbf{z}) \bigr) 
        =\mathrm{KL}\bigl( \mathcal{N}(\bar \mu, \bar\Sigma) \,\|\, \mathcal{N}(0, I) \bigr)
        \\
        =\frac{1}{2} \sum_{j=1}^d \left[ \bar\sigma_j^2 + \bar\mu_j^2 - 1 - \log \bar\sigma_j^2\right];
\end{multline}
where we define covariance diagonal matrix of the approximated posterior as $\bar \Sigma = \mathrm{diag}(\bar \sigma^2)$, and $\bar \mu$ and $\bar \sigma^2$ are given in \eqref{eq:mu_bar2} and \eqref{eq:sigma2_bar2}, respectively.

Furthermore, one behavior we aim to prevent is the collapse of the variances of individual posteriors $(\sigma^{(i)})^2$ to zero. That eliminates the stochastic behavior of the Encoder, causing the VAE to behave like an AE. In such a scenario, the VAE would be limited to generalizing: to generate data not part of its training data \cite{hinton2006:reddim}. Thus, we introduce a regularization term that penalizes individual variances that become close to zero, preventing such individual posteriors from becoming Dirac deltas. This regularization term is:
\begin{multline}
\label{eq:KL_I}
\mathrm{KL}_I \bigl( q_\phi(\mathbf{z}|\mathbf{x}) \, \| \, p_\lambda(\mathbf{z}) \bigr) 
\\
= \sum_{i:\x_i \in \D} \mathrm{KL}\bigl( q_{\phi}(\z^{(i)}|\x^{(i)}) \, \| \, \mathcal{N}(\mu^{(i)}, I) \bigr) \\
    = \sum_{i:\x_i \in \D}  \sum_{j=1}^d \left[(\sigma^{(i)}_j)^2 - 1 - \log (\sigma^{(i)}_j)^2\right].
\end{multline}
Note that KL \eqref{eq:KL_I} does not penalize the individual means.

It is well known that metrics based on pixel-to-pixel comparisons are not the most appropriate for evaluating the quality of data produced by generating models. We will tolerate significant reconstruction errors as long as the reconstruction looks realistic. In the example we use, this means that we prefer to generate images where the hairstyle seems realistic, even if the positions of the curls in the reconstruction do not match the input image, rather than over-smoothing regions of the hair (as is common in a typical VAE). GAN-type models have been relatively successful in generating realistic images \cite{creswell2018generative}, despite suffering from the problem of modes collapse. Similar to VAE-GAN, which trains the VAE based on a GAN-type scheme \cite{larsen2016autoencoding}, we also use a GAN training scheme. In particular, we use a PatchGAN strategy \cite{li2016precomputed,isola2017image,Chang_2019_ICCV}, which, instead of giving a reality measurement rating for the entire image, generates a two-dimensional array where each element evaluates a support region in the generated image or real.

Therefore, the likelihood term includes two metrics: a pixel-to-pixel measure that enforces the restoration to be similar to the input data and a general evaluation grade of the image realism.
For the first term we choose the L1 norm known for being robust to errors in the exact reconstruction of structures' positions in high-frequency texture areas. This is equivalent to said that the error follows a Laplace distribution.
For the second term, we use a Convolutional Neural Network (Discriminator) that, given the restored image $\hat \x$, of size (H,W), produces an output $\r$, of size $(h/8, w/8)$, where each entry reflects the realism degree of its support region. The discriminator is represented by the conditional $\r \sim p_\gamma (\r | \x)$ where $\r \in [0,1]^{h/8 \times w/8}$, and $\gamma$ are the parameters of the neural network. Then the log--likelihood takes the form
\begin{multline}
\label{eq:nlog_H}
\mathbb{E}_{q_\lambda(\mathbf{z}|\mathbf{x})} \left[ \log p_\theta(\mathbf{x}|\mathbf{z}) \right] 
 =  \beta_3 \|\x - \hat \x \| -\beta_4 \mathrm{H}\left[ \1 , p_\gamma(\r|\hat \x )\right]
\end{multline}
where $\mathrm{H}$ is the cross-entropy, $(\beta_3, \beta_4)$ are scale parameters, $\1$ is a vector with all its entries equal 1 which size depends on the context. Thus the regularized ELBO takes the form:
\begin{multline}
\label{eq:ELBO_reg}
\mathrm{ELBO}_{reg}(\phi, \theta) 
= 
\beta_1 \, \mathrm{KL}_G \bigl( q_\phi(\mathbf{z}|\mathbf{x}) \, \| \, p_\lambda(\mathbf{z}) \bigr)  \\
+ \beta_2 \, \mathrm{KL}_I \bigl( q_\phi(\mathbf{z}|\mathbf{x}) \, \| \, p_\lambda(\mathbf{z}) \bigr)  \\ 
+\beta_3 \|\x - \hat \x \| -\beta_4 \mathrm{H}\left[ \1 , p_\gamma(\r|\hat \x)\right]
\end{multline}
This loss encourages the generator to produce realistic predictions $\hat \x$. To train the discriminator, we minimize the loss
\begin{equation}
\label{eq:loss_D} 
L(\gamma) =  -\beta_4 \left( \mathrm{H}\left[ \1 , p_\gamma(\r| \x )\right] + \mathrm{H}\left[ \0 , p_\gamma(\r|\hat \x )\right] \right)
\end{equation}

\begin{figure}
    \centering
    \includegraphics[width=\linewidth]{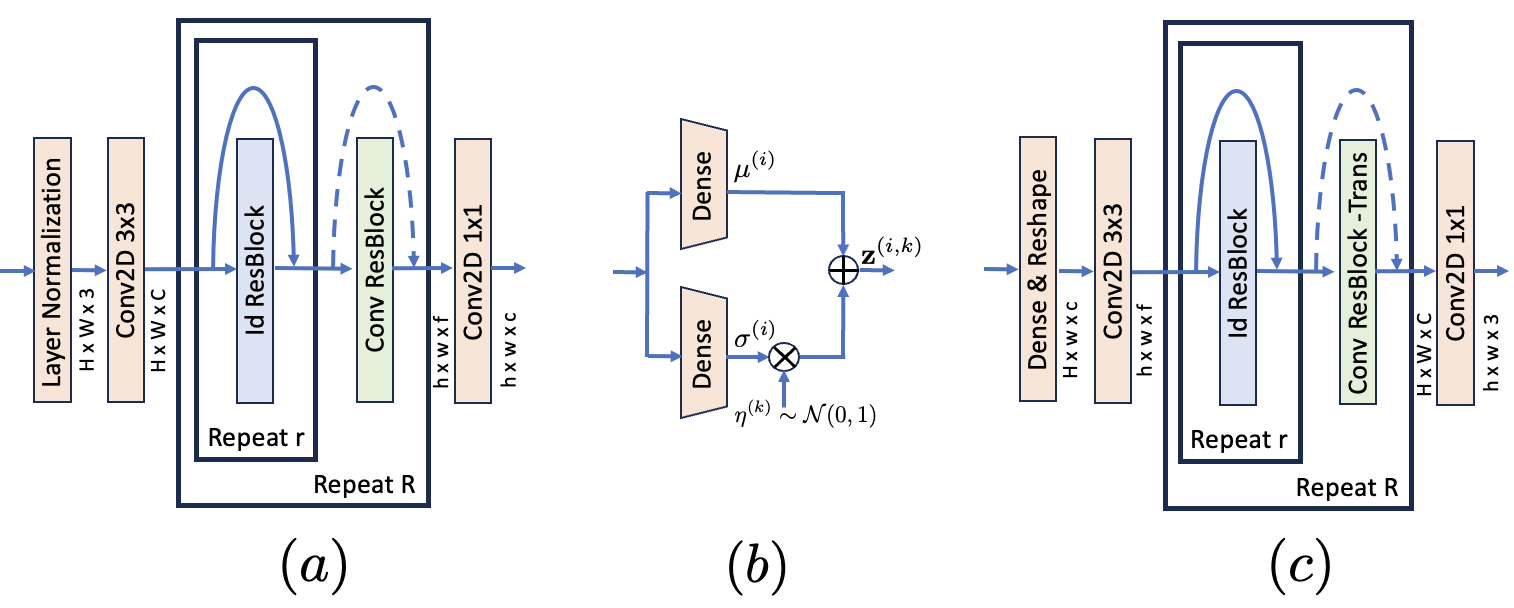}
    \caption{Residual-VAE components. (a) Residual encoder, (b) Gaussian sampler, and (c) residual decoder. In the presented experiments, we use an architecture with six residuals blocks ($R=6$) with two identity blocks each one ($r=2$).}
    \label{fig:ResidualVAE}
\end{figure}

In all the conducted experiments, we kept the weight parameters: $(\beta_1, \beta_2, \beta_3,\beta_4) = (1, 0.5, 5000, 100)$. For computational efficiency purposes, we implement the training of both the VAE (generator) at the discriminator, using a stochastic descent strategy with batch size equal to $b=20$. Note that as the batch size $b$ becomes larger, the estimation of ELBO terms [Eqs. \eqref{eq:KL_G}, \eqref{eq:KL_I} and \eqref{eq:nlog_H}] becomes more precise. Therefore, avoiding using batches close to one is important to ensure stable convergence of the stochastic gradient. Algorithm 1 summarizes training details.
\begin{algorithm}
\caption{VAE-GAN Training}
\label{alg:VAEGAN}
Given the input batch $\x$ with dimensions $(b,h,w,f)$.
\begin{algorithmic}[1]
\Procedure{TrainStep}{}
\State $\z \sim q_\phi(\z | \x)$           \Comment{Encode}
\State $\hat \x \sim p_\theta(\x | \z)$    \Comment{Decode}
\State $\r \sim p_\gamma(\r | \hat \x)$    \Comment{Realism scores}
\State Take a gradient descent step on $ \nabla_{\phi,\theta} ELBO_{reg}(\phi,\theta)$.
\State Take a gradient descent step on $ \nabla_{\gamma} L(\gamma)$.
\EndProcedure
\end{algorithmic}
\end{algorithm}

\subsection{Network architecture}
\label{ssec:architecture}

Fig. \ref{fig:ResidualVAE} depicts the components in our VAE implementation. 
The architecture of a VAE adheres to a foundational framework, depicted in Fig. \ref{fig:vae}. Within this framework, the two fundamental components of an AE persist the Encoder (ENC) and the Decoder (DEC). However, the hallmark that sets the VAE apart is introducing a distinctive element: the 
sampler. We based our ResidualVAE implementation on ResNetV2 \cite{he2016identity}. In particular, the Identity Residual Block (Id-ResBlock), Convolutional Residual Blocks (Conv-ResBlock), and  Convolutional Residual Transposed Blocks (Conv-ResBlock-T). 
The "Id-ResBlock" maintains input and output dimensions, the "Conv-ResBlock" reduces dimensions through a stride in the convolutional layer, and
"Conv-ResBlock-T" extends the spatial dimensions through an interpolation performed by a transposed convolution layer.

\section{Experiments}
\label{sec:experiments}

We show experiments designed to demonstrate that our proposal preserves the essential characteristics of VAEs while generating high-quality data. That is, the generated image is similar to the image that generated the latent variable; small variations in the latent variables introduce slight variations in the generated faces, and the interpolation of latent vectors produces realistic data with smooth transitions.

In the presented cases we use the Adam optimization method for training the Residual-VAE and the Discriminator, we set the initial learning rate equal to $10^{-4}$. To demonstrate our proposal's performance, we generate faces of $256 \times 256$ pixels in RGB using the CelebA-HD database \cite{karras2018progressive}. After the alphanumeric ordering of files, this database consists of 30,000 images that we split in the first 24,000 for training and the reminder 6000 for testing.

\begin{figure}[!ht]
\centering
\includegraphics[width=\linewidth]{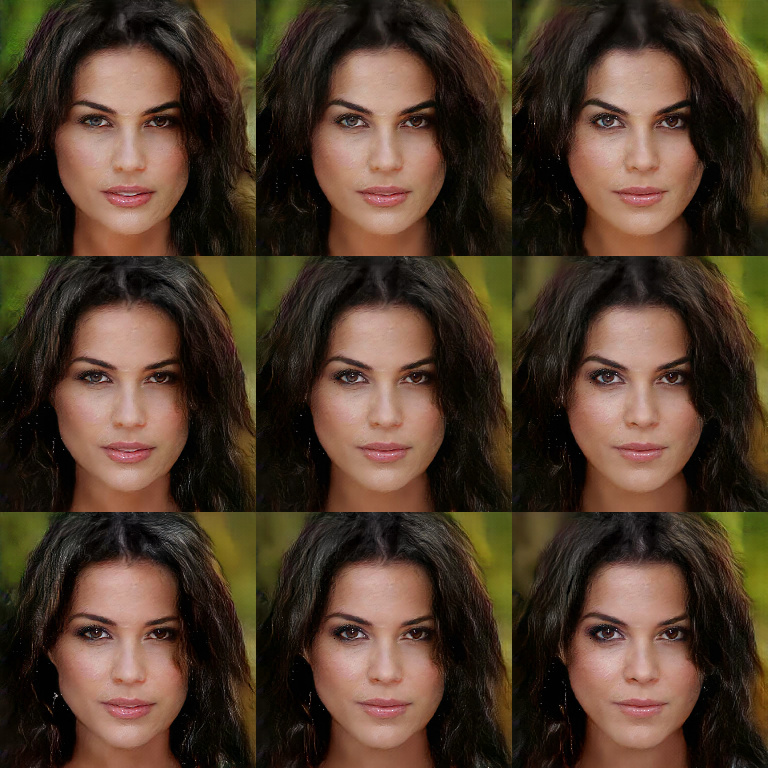}
\caption{Generated variants with the VAE using the proposed training scheme. Varying $z_0$ and $z_1$ according $\delta_k \times \sigma^{(i)}$ for $\delta=[-20,0,20]$.} 
\label{fig:variants}
\end{figure}

Our first experiment demonstrates the generalization capacity of our proposal. We generate variants of a random face ${\bf x}^\prime$, from the test set. We obtain the variations by sampling the posterior approximation:
$
\z^\prime \sim q_\phi({\bf z}| {\x=\bf x}^\prime).
$
Next, we select two entries from the latent vector: $a, b \in [1, m]$. Then, with all inputs of $\z^\prime$ fixed except $\z_a^\prime$ and $\z_b^\prime$, we generate images that sample the probability ($\hat{\bf x} \sim p_\theta ( {\x | \z^\prime})$ varying these entries for combinations of $\z_a^\prime = \mu_a^{(i)} + \delta_k \sigma_a^{(i)}$ with $\delta=[-20, 0, 20]$  and likewise for $\z_b^ \prime$. Fig. \ref{fig:variants} shows the generated variants. As we can see, the generated faces conserve a large part of the central face. In the illustrated example, the variables $\z_0$ and $\z_2$ correspond to the horizontal and vertical axes, respectively.

\begin{figure*}[!ht]
    \centering
    \includegraphics[width=\textwidth]{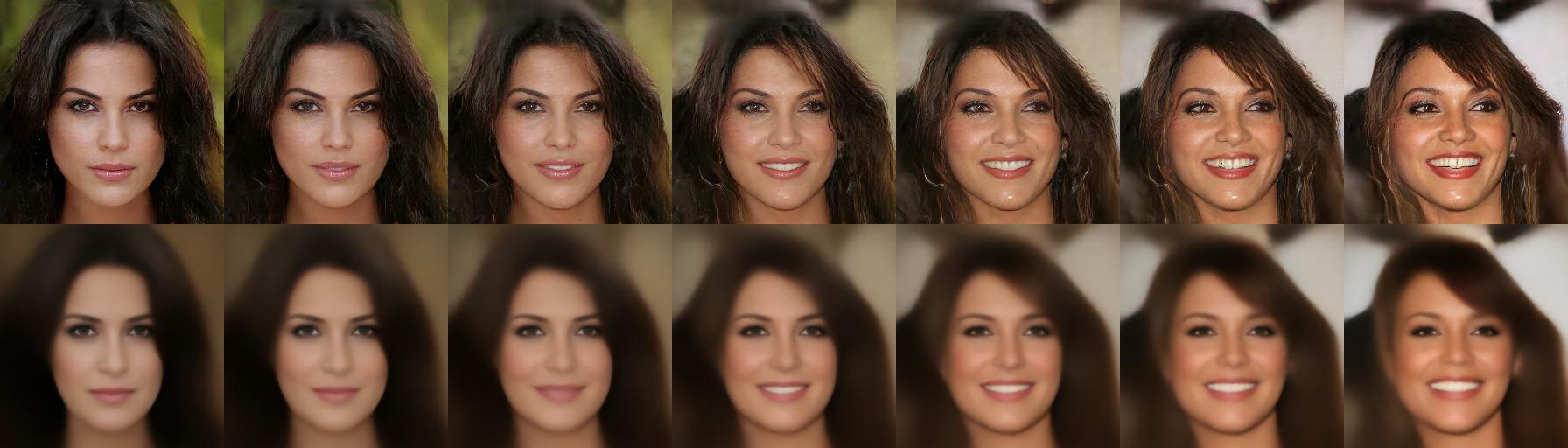}
    \caption{Transitions generated with VAEs by convex combinations of latent variables: $\hat \x \sim p_\theta(\x | \z = \alpha \,\z^{(i)} + (1-\alpha) \,\z^{(j)} ) $, for $\alpha \in [0,1]$. First row, VAE trained with the proposed scheme. Second row, the same VAE model trained with the standard training strategy: $\beta =[1,5000]$ in \eqref{eq:ELBO_KL}-\eqref{eq:ELBO_l}, and $L_1$ norm for the negative log-likelihood.}
    \label{fig:transition}
\end{figure*}

Another interesting feature of VAEs is the ability to generate transitions between two faces, where the convex combination of two latent variables corresponds to a face with characteristics smoothly transitioning from one face to another. Fig. \ref{fig:transition} depicts generated transitions. Each row corresponds to the transition between two latent variables ${\bf z}_1$ and ${\bf z}_2$, corresponding to randomly selected images in the test dataset. The columns correspond to
$
{\bf z}_k^\prime = (1-\alpha_k) \, {\bf z}_1 + \alpha_k \,{\bf z}_2
$,  with $\alpha_k = \frac{k}{(n-1)}$; for $k=0,1,\ldots,5$.

\begin{figure}
    \centering
    \includegraphics[width=0.75\linewidth]{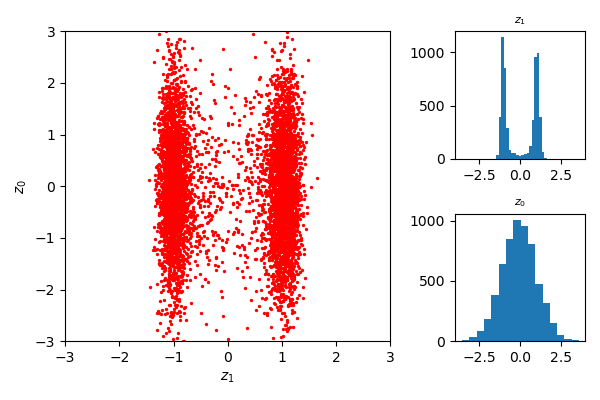}
    \caption{Left, the join histogram of the latent variables $z_0$ and $z_1$ for the test dataset. On the right are the respective marginal histograms.}
    \label{fig:joint_histo}
\end{figure}

Finally, we show in Fig. \ref{fig:joint_histo} the joint histogram of two variables with typical estimated distributions. The histograms correspond to the latent variables $\z_0$ and $\z_1$ for 400 test data points; on the right the marginal histograms. It is notable that in our formulation, 198 of 512 latent entries presented a bimodal histogram (with modes at -1 and 1). In the case of beta-VAE with $\beta_1 < \beta_2$, the collapsed histograms are monomodal and approach a Dirac delta, as noted in Ref. \cite{hinton2006:reddim}. A more in-depth analysis of the use of these collapsed variables is beyond the scope of this article.

\section{Conclusions}
\label{sec:conclusions}

Our proposal substantially improves the reported results for standard VAE models while preserves the essential generative characteristics of the VAE. Although more complex variations of VAEs have been proposed (e.g., Hierarchical VAEs \cite{havtorn2021hierarchical} and VQ-VAEs \cite{razavi2019generating}), standard VAEs continue to be of great interest due to their elegant theoretical formulation and the promise of generating complex data from a simple distribution. Their main drawback is the posterior collapse problem, resulting in smoothed data. In this work, we propose a solution to reduce these effects. Through experiments, we have demonstrated that reinterpreting the posterior as a mixture of Gaussians leads to a variant of the ELBO with a marginal computational cost. The list of our contributions in this work is as follows. First, an ELBO that
reduces the collapse of the posterior towards the anterior; observed as the generation of very similar blurry images. Our ELBO can be seen as an opposite version of Beta-VAE that preferentially weights quality reconstructions (with $\beta<1$), which is regularized to prevent individual posteriors from collapsing to a Dirac delta. Additionally, we use a VAE-PatchGAN training scheme that promotes realistic data through a discriminator that evaluates the generated data's realism and reduces the metric's contribution that compares pixel by pixel. Finally, we redesigned the encoder and decoder architecture to be of the Residual type to increase the encoding and decoding capabilities. Since the main contributions affect the training process, our results must be compared directly with those of a standard VAE or Beta-VAE. Qualitative comparisons of the results presented against those reported in the classic VAE formulations show the superior performance of our proposal. The code that implements the proposal is available in \url{https://github.com/marianorivera/How2TrainUrVAE}.

\vspace{2mm}
\noindent {\bf Acknowledges.} Work supported in part by CONAHCYT, Mexico (Grant CB-A1-43858). The author thanks to the anonymous reviewers and L. M. Nunez for their useful comments to improve the quality of this paper.
\vspace{-3mm}





\end{document}